%% file: main.tex
\documentclass[runningheads]{llncs}

\usepackage{eccv}

\usepackage{eccvabbrv}

\usepackage{graphicx}
\usepackage{booktabs}
\usepackage{multirow}
\usepackage{tabularx}
\usepackage[accsupp]{axessibility}  

\usepackage{hyperref}

\usepackage{orcidlink}
\newcommand\blfootnote[1]{%
  \begingroup
  \begin{NoHyper}%
  \renewcommand\thefootnote{}\footnote{#1}%
  \addtocounter{footnote}{-1}%
  \end{NoHyper}%
  \endgroup
}

\begin{document}

\title{RepVF: A Unified Vector Fields Representation for Multi-task 3D Perception}



\titlerunning{RepVF: Unified Multi-task 3D Perception}


\author{
Chunliang Li\inst{1}$^*$\orcidlink{0009-0002-9756-4488} \and
Wencheng Han\inst{2}$^*$\orcidlink{0009-0005-2358-6969} \and 
Junbo Yin\inst{1}\orcidlink{0009-0000-1127-3686} \and \\ 
Sanyuan Zhao\inst{1}\orcidlink{0000-0001-9386-9677}$^\dagger$
 \and
Jianbing Shen\inst{2}\orcidlink{0000-0003-1883-2086}, 
}
\authorrunning{C. Li et al.} 
\institute{School of Computer Science, Beijing Institute of Technology \\ 
\and SKL-IOTSC, Computer and Information Science, University of Macau, China \\
\email{\{jbji, sanyuanzhao\}@bit.edu.cn \\ \{wencheng256, yinjunbocn\}@gmail.com \\ jianbingshen@um.edu.mo}
}

\maketitle

\begin{abstract}

Concurrent processing of multiple autonomous driving 3D perception tasks within the same spatiotemporal scene poses a significant challenge, in particular due to the computational inefficiencies and feature competition between tasks when using traditional multi-task learning approaches. 
This paper addresses these issues by proposing a novel unified representation, RepVF, which harmonizes the representation of various perception tasks such as 3D object detection and 3D lane detection within a single framework. RepVF characterizes the structure of different targets in the scene through a vector field, enabling a single-head, multi-task learning model that significantly reduces computational redundancy and feature competition.
Building upon RepVF, we introduce RFTR, a network designed to exploit the inherent connections between different tasks by utilizing a hierarchical structure of queries that implicitly model the relationships both between and within tasks. This approach eliminates the need for task-specific heads and parameters, fundamentally reducing the conflicts inherent in traditional multi-task learning paradigms.
We validate our approach by combining labels from the OpenLane dataset with the Waymo Open dataset. Our work presents a significant advancement in the efficiency and effectiveness of multi-task perception in autonomous driving, offering a new perspective on handling multiple 3D perception tasks synchronously and in parallel. The code will be available at: \url{https://github.com/jbji/RepVF}.

\end{abstract}

\blfootnote{$^{*}$Equal contribution. $^{\dagger}$Corresponding author. 
This work was supported in part by the FDCT grants
0102/2023/RIA2, 0154/2022/A3,
and 001/2024/SKL, the MYRG-CRG2022-00013-IOTSC-ICI grant and the SRG2022-00023-IOTSC grant.
}

\input{sec/1_introduction}
\input{sec/2_related_work}
\input{sec/3_RepVF}

\input{sec/4_RFTR}

\input{sec/5_experiments}

\section{Conclusion}
In this paper, we propose \textit{RepVF}, a novel unified vector fields representation that characterizes the geometric and semantic structure of different 3D perception targets in the scene. Built upon \textit{RepVF}, our proposed RFTR exploits the intrinsic connections between tasks, mitigates multi-task gradient imbalance and feature competition, and improves model convergence and multi-task expression. Our work not only contributes a novel single-head perspective to the multi-task learning paradigm in autonomous driving but also sets a promising direction for future research in efficient and effective 3D perception task handling.


%
%
\bibliographystyle{splncs04}
\bibliography{main}

\end{document}

%% file: sec/1_introduction.tex
\section{Introduction}

\begin{figure}[tb]
  \centering
  \resizebox{0.97\textwidth}{!}{
  \begin{subfigure}{0.49\linewidth}
    \includegraphics[width=\textwidth]{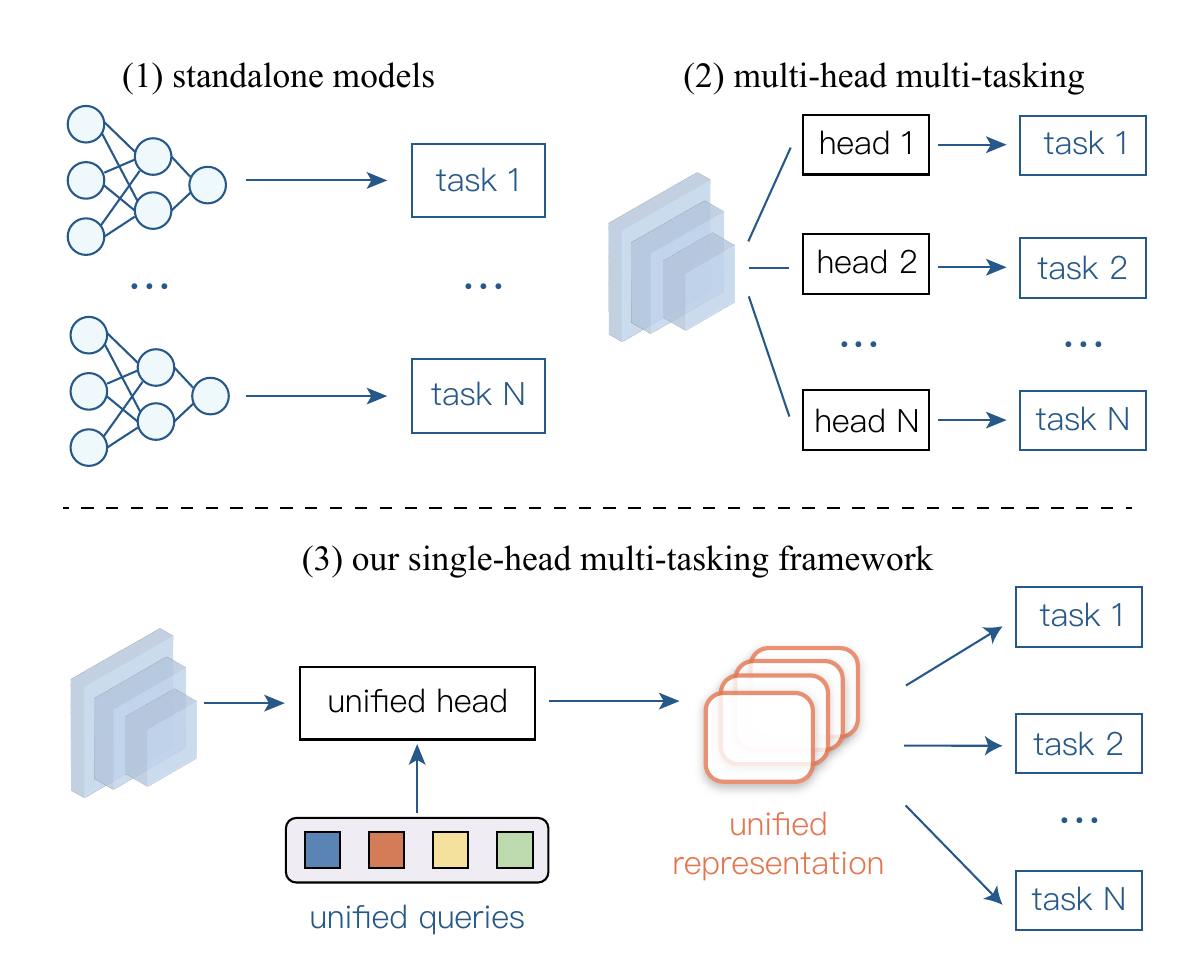}
    \caption{Multi-tasking paradigm comparison.}
    \label{fig:paradigm-a}
  \end{subfigure}
  \hfill
  \begin{subfigure}{0.47\linewidth}
    \includegraphics[width=\textwidth]{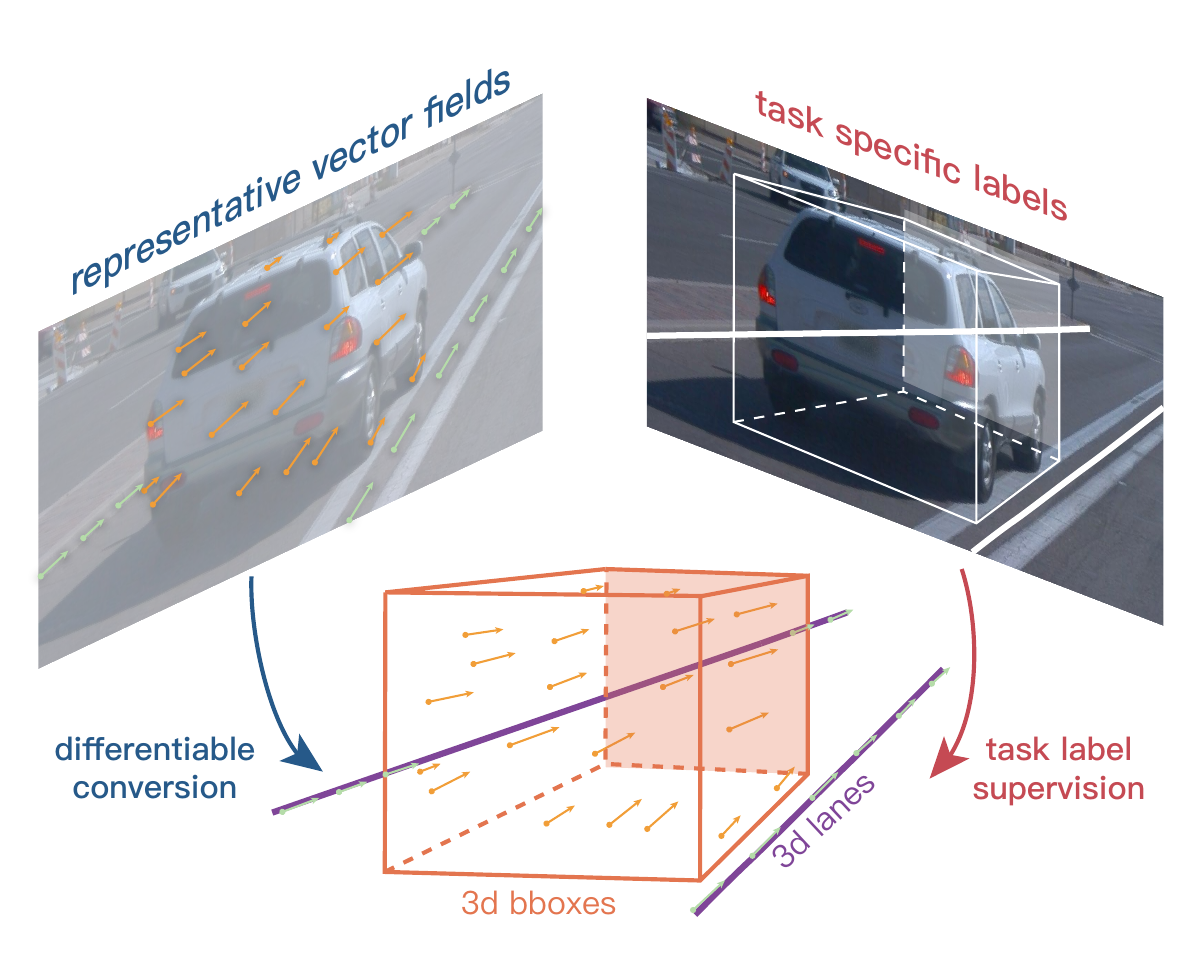}
    \caption{\textit{RepVF}, conversion and supervision.}
    \label{fig:paradigm-b}
  \end{subfigure}
  }
  \caption{\textit{RepVF} is proposed as a unified representation for 3D perception to achieve single-head multi-tasking. It consists of a set of vector fields learned to represent road elements, and is trained utilizing existing labels through differentiable conversions. The proposed single-head multi-tasking paradigm reduces task conflict and competition.}
  \label{fig:paradigm}
\end{figure}

In autonomous driving, multiple 3D perception tasks often need to be processed synchronously, in real-time, and in parallel. These tasks, which exist within the same spatiotemporal scene, are often complementary to each other in both geometric and semantic aspects\cite{huangFULLERUnifiedMultimodality2023}. For example, 3D object detection\cite{geigerAreWeReady2012} and 3D lane detection\cite{garnett3DLaneNetEndtoEnd3D2019a} define foreground objects and background lanes within the same 3D scene. Vehicles usually travel following the guiding semantics of lanes that do not spatially overlap and are partially obscured by other vehicles. Some previous industrial solutions~\cite{NVIDIADRIVESolutions, MobileyeCES2024} utilize a multi-model paradigm for multi-task perceptions, as depicted in \Cref{fig:paradigm-a}(1). These separate models cannot share common features across different tasks, resulting in computational waste. Furthermore, this paradigm does not take advantage of the inherent connections between different tasks. Some works~\cite{liBEVFormerLearningBird2022, liuPETRv2UnifiedFramework, huangFULLERUnifiedMultimodality2023, liuBEVFusionMultiTaskMultiSensor2023} attempt to overcome this limitation by using a common backbone and separate head networks for each task, as shown in \Cref{fig:paradigm-a}(2). This approach can considerably reduce the computational cost when handling a large number of tasks, but it might lead to a task balancing issue. Since different task-aware heads focus on various aspects of the scene, they may compete for the backbone features, resulting in instability during training~\cite{huangFULLERUnifiedMultimodality2023, liuBEVFusionMultiTaskMultiSensor2023}.

The competition is attributed to the variation in representation between different tasks. For example, the 3D detection task~\cite{geigerAreWeReady2012, wangDETR3D3DObject2022, liuPETRv2UnifiedFramework, liuPETRPositionEmbedding2022a, liBEVFormerLearningBird2022, Li_Yin_Shi_Li_Yang_Shen_2023, Wang_Yin_Li_Frossard_Yang_Shen_2023} 
guides the head network to predict 3D bounding boxes, whereas the lane detection task\cite{garnett3DLaneNetEndtoEnd3D2019a, chenPersFormer3DLane2022a} necessitates the corresponding head to regress lane shapes. Specifically, 3D bounding boxes describe a cubic area in 3D space, representing different category foregrounds using position, size, and orientation. In contrast, 3D lanes are represented using anchor\cite{garnett3DLaneNetEndtoEnd3D2019a, chenPersFormer3DLane2022a, huangAnchor3DLaneLearningRegress2023a} or parametric\cite{hanDecouplingCurveModeling2023} methods to describe one or more 1-D lines. These representations focus on different aspects of the scenes, which ultimately leads to competition.

In this paper, we propose a unified representation, \textit{RepVF} (Representative Vector Fields), for perception tasks in autonomous driving. Using this representation, we design a single-headed, unified model that can perform tasks simultaneously.
As illustrated in \cref{fig:paradigm-b}, \textit{RepVF} represents 3D vector fields that each assigns vectors to spatial locations, denoted as  $(S,\mathcal{F}(S)) \subseteq \mathbb{R}^{3+d}$ in \cref{sec:repvf-definition}.
Sub-vector fields representing different targets in the 3D perception scene progressively adhere to their spatial extent, accurately characterizing the anisotropic structure in their respective spatial vicinity.
\textit{RepVF} can be differentiably transformed into task-specific elements, hence requiring no special supervision and can utilize existing labels. 
With \textit{RepVF}, our network only processes one single type of fundamental perception element. As shown in \cref{fig:paradigm-a}, unlike the traditional multi-task approach that employs multiple task heads\cite{yin2020unified, liangBEVFusionSimpleRobust2022, huangFULLERUnifiedMultimodality2023,liBEVFormerLearningBird2022, liuPETRv2UnifiedFramework}, we follow an unprecedented single-head multi-task paradigm. The single-head multi-task eliminates task-specific parameters and obviates the need to explicitly model task interactions. 

\textit{RepVF} representation serve as a geometrical, cross-scale 3D perception representation alternative. Unlike perception elements capable of handling only single-scale tasks\cite{geigerAreWeReady2012, wangDETR3D3DObject2022, garnett3DLaneNetEndtoEnd3D2019a, chenPersFormer3DLane2022a}, it breaks down the scene into smaller perceptual units while retaining the capability for target-level perception at larger scales. 
Based on the \textit{RepVF} representation, we further propose RFTR to leverage this relationship, utilizing queries with a hierarchical structure that implicitly models the relationships both between and within different tasks. 
Following the single-head multi-task architecture, with only shared parameters across tasks, RFTR does not depend on existing multi-task optimization strategies~\cite{kendallMultiTaskLearningUsing2018, senerMultiTaskLearningMultiObjective2018, liuEndtoEndMultiTaskLearning2019, liuImpartialMultitaskLearning2020, yuGradientSurgeryMultiTask2020}. The gradient discrepancies between tasks across different training iterations have been mitigated,
fundamentally reducing the competition and conflict inherent in multi-tasking 3D perception.
By employing \textit{RepVF} to replace the fundamental elements for different tasks, and by combining 3D lane labels from the OpenLane dataset\cite{chenPersFormer3DLane2022a} with the Waymo Open dataset\cite{sunScalabilityPerceptionAutonomous} for simultaneous multi-task training and inference, 
our approach achieves comparable or even better multi-task performance than single-task counterparts with only one unified model.

In summary, we introduce \textit{RepVF} as a unified representation for multiple perception tasks in autonomous driving, effectively reducing the competition among different task representations during training and enhancing model convergence. We further develop a single-head multi-task framework RFTR based on \textit{RepVF}, fully exploiting the intrinsic connections among different tasks within a scene, improving feature interaction with queries, and enabling gradient-balanced multi-task training. Our approach achieves impressive performance across two key tasks in autonomous driving—3D object detection and 3D lane detection—showcasing the advantages of a unified task representation in the context of autonomous driving applications.

%% file: sec/2_related_work.tex
\section{Related Work}

\subsection{Camera-based 3D Perception in Autonomous Driving}
3D object detection\cite{geigerAreWeReady2012, chenMonocular3DObject2016, wangDETR3D3DObject2022, yinSemisupervised3DObject2022, chengLanguageGuided3DObject2023a} and 3D lane detection\cite{garnett3DLaneNetEndtoEnd3D2019a, guoGenLaneNetGeneralizedScalable2020} are two common 3D perception tasks, and previous approaches usually perform the two tasks independently. 3D object detection has been extensively studied either in model architectures\cite{chenMonocular3DObject2016, wangFCOS3DFullyConvolutional2021, liuPETRPositionEmbedding2022a, liuPETRv2UnifiedFramework} or feature representations\cite{roddickOrthographicFeatureTransform2018, zhouIafaInstanceawareFeature2020, liBEVFormerLearningBird2022, huangBEVDet4DExploitTemporal2022,liUnifyingVoxelbasedRepresentation2022, readingCategoricalDepthDistribution2021, liDiv2xLearningDomaininvariant2024, yinIsfusionInstancesceneCollaborative2024}. Early methods\cite{mousavian3DBoundingBox2017, kehlSSD6DMakingRGBBased2017, kuMonocular3DObject2019} predict 3D bounding boxes using 2D detectors, and Mono3D\cite{chenMonocular3DObject2016} is one of the first methods directly make predictions from 3D representations. Motivated by DETR\cite{garnett3DLaneNetEndtoEnd3D2019a}, DETR3D\cite{wangDETR3D3DObject2022} uses sparse 3D object centers as queries, and PETR\cite{liuPETRPositionEmbedding2022a} further enhances its feature representations by embedding 3D positions. Some recent approaches \cite{liBEVFormerLearningBird2022, huangBEVDet4DExploitTemporal2022, liuPETRv2UnifiedFramework, wangExploringObjectCentricTemporal2023a} also focus on incorporating temporal information into 3D object detection, but they all follow the transformer-based\cite{vaswaniAttentionAllYou2017a} paradigm introduced by DETR3D\cite{wangDETR3D3DObject2022}.

Only recently have DETR-based methods been introduced for 3D lane detection \cite{baiCurveFormer3DLane2023a, luoLATR3DLane2023a, liuPETRv2UnifiedFramework, wu2023topomlp, wu20231st}. Previous methods \cite{garnett3DLaneNetEndtoEnd3D2019a,guoGenLaneNetGeneralizedScalable2020, chenPersFormer3DLane2022a, wangBEVLaneDetEfficient3D2023, liuLearningPredict3D2022} prefer to project image features into BEV space using Inverse Perspective Mapping (IPM), but this introduces prohibitive height prediction error. Some work \cite{yanONCE3DLanesBuildingMonocular2022} avoids IPM via depth estimation, and Anchor-3DLane \cite{huangAnchor3DLaneLearningRegress2023a} reverses this projection process. Following DETR\cite{wangAnchorDETRQuery2022,baiCurveFormer3DLane2023a, liuPETRv2UnifiedFramework} introduce 3d lane anchors as queries, and \cite{luoLATR3DLane2023a} adopts finer lane prior queries and is end-to-end. 
With the advances of DETR\cite{wangAnchorDETRQuery2022, wangDETR3D3DObject2022} networks in both tasks\cite{liuPETRPositionEmbedding2022a, liBEVFormerLearningBird2022, baiCurveFormer3DLane2023a, luoLATR3DLane2023a}, it is architecturally ready to unify the two tasks without standalone models.

\subsection{Multi-task 3D Perception}

The problem of multi-tasking in autonomous driving perception is still in its infancy. Recent advances \cite{philionLiftSplatShoot2020, zhouJoint3dInstance2020,  liBEVFormerLearningBird2022, xieBEVMultiCameraJoint2022, liuPETRPositionEmbedding2022a, liuPETRv2UnifiedFramework, huangFULLERUnifiedMultimodality2023, liuBEVFusionMultiTaskMultiSensor2023} elaborate on universal perception features and multiple tasks are handled with different heads. Moreover, there is no consensus on the required tasks. For example, MMF\cite{liangMultiTaskMultiSensorFusion2019} addresses both depth completion and object detection issues, while \cite{ huangFULLERUnifiedMultimodality2023, liBEVFormerLearningBird2022, liuPETRv2UnifiedFramework, liuBEVFusionMultiTaskMultiSensor2023} performs object detection and map segmentation, with \cite{liuPETRv2UnifiedFramework} performing additional 3D lane detection separately. However, most of these studies only use multiple tasks to demonstrate the universality of their learned features, and the tasks are trained and evaluated separately\cite{liBEVFormerLearningBird2022, liuPETRv2UnifiedFramework} rather than simultaneously\cite{liuBEVFusionMultiTaskMultiSensor2023, huangFULLERUnifiedMultimodality2023}, not to mention dataset discrepancies.

In addition, multitask learning (MTL) typically involves task conflict. Previous MTL studies\cite{vandenhendeMultiTaskLearningDense2022a} addresses this either through parameter sharing \cite{nevenFastSceneUnderstanding2017, teichmannMultiNetRealtimeJoint2018a, misraCrossStitchNetworksMultiTask2016, xuPADNetMultiTasksGuided2018, ruderLatentMultiTaskArchitecture2019, guoLearningBranchMultiTask2020} or optimization\cite{kendallMultiTaskLearningUsing2018, senerMultiTaskLearningMultiObjective2018, chenGradNormGradientNormalization2018a, liuEndtoEndMultiTaskLearning2019, liuImpartialMultitaskLearning2020, yuGradientSurgeryMultiTask2020}. 
Parameter sharing approaches are conceptually straightforward, and are further classified into hard-sharing (with shared and task-specific parameters) and soft-sharing (with a cross-task talk mechanism).
Optimization-based approaches, \eg IMTL\cite{liuImpartialMultitaskLearning2020} and DWA\cite{liuEndtoEndMultiTaskLearning2019}, attribute MTL imbalance to the gradient and loss imbalance and explore optimization calibration methods. Taking advantage of these advances, FULLER\cite{huangFULLERUnifiedMultimodality2023} stands as the pioneering work in autonomous driving perception that analyzes the performance of modern optimization-based MTL calibration methods\cite{chenGradNormGradientNormalization2018a, liuImpartialMultitaskLearning2020, liuEndtoEndMultiTaskLearning2019}. With a unified task representation, we follow a different paradigm from previous ones, that eliminate task-specific heads and mitigate the gradient imbalance. Our approach can be seen as a hard-sharing technique without task-specific parameters and does not require calibration.

%% file: sec/3_RepVF.tex
\section{Representative Vector Fields}
We begin by revisiting the representations of two common 3D perception tasks: 3D object detection\cite{geigerAreWeReady2012, sunScalabilityPerceptionAutonomous, caesarNuScenesMultimodalDataset2020} and 3D lane detection\cite{chenPersFormer3DLane2022a, yanONCE3DLanesBuildingMonocular2022}.  
By exploring how task-specific representations have evolved into task-specific designs\cite{chenMonocular3DObject2016, roddickOrthographicFeatureTransform2018, wangDETR3D3DObject2022, liuPETRPositionEmbedding2022a, garnett3DLaneNetEndtoEnd3D2019a, chenPersFormer3DLane2022a, liuPETRv2UnifiedFramework,baiCurveFormer3DLane2023a, luoLATR3DLane2023a}
, we propose the task-agnostic \textit{RepVF} (Representative Vector Fields), which arise from the geometric commonality in the tasks and effectively reduce task competition via unified representations.

\subsection{Task-specific Representations and Single-tasking}

\subsubsection{Representation Formulations.} Task-specific representations vary due to the geometric features of the task targets, which differ in dimensions and scales. 
As a key task for understanding 3D space, 3D object detection deals with objects irregularly shaped and scaled from one to several metres, defined by coarse outer boundaries as 3D bounding boxes \cite{geigerAreWeReady2012} , formally a 7-d coordinate $B=(x,y,z,l,w,h,\theta)$ encoding object center $(x,y,z)$, box dimensions $(l,w,h)$ and heading $\theta$. Conversely, due to their linear shape and ability to cover long distances, 3d lanes are not described by boundaries but by direct geometries, typically ordered sequences of $N_L$ 3D point coordinates \cite{chenPersFormer3DLane2022a}: 
\begin{equation}
    L=[(x_1,y_1,z_1),(x_2,y_2,z_2),\dots,(x_{N_l},z_{N_l},z_{N_l})]
\end{equation}
This may look familiar to an unordered point cloud, but they are ordered along the direction of the road. In addition, 3D perception problems typically involve classifications that require category scores $\mathcal{C}_i$, varying from 1, 3 or 23 \cite{geigerAreWeReady2012, sunScalabilityPerceptionAutonomous, caesarNuScenesMultimodalDataset2020} classes (3D object) to 21 or 22 \cite{chenPersFormer3DLane2022a} classes (3D lane).

\subsubsection{Evolving into Task-specific Designs.}

Pioneering 3D object detection from RGB images is proposal based, \eg Mono3D\cite{chenMonocular3DObject2016} 
reduces proposal space dimension of each class to $(x,y,z,\theta,t)$ with representative templates, and only predicts $\Delta_{\text{pos}}$;
OFT-Net\cite{roddickOrthographicFeatureTransform2018} predicts offsets of all bounding box parameters relative to the top-view grid, \ie $\Delta_{\text{pos}}, \Delta_{\text{dim}}, \Delta_{\text{ang}}$. 
Taking advantage of the center coordinates, DETR-based methods\cite{wangDETR3D3DObject2022,liuPETRPositionEmbedding2022a} regress $\Delta_{\text{pos}}$ along with bounding box parameters. However, constrained by the limited spatial extent of center points, extracted features around them capture only local spatial areas\cite{zhuDeformableDETRDeformable2021d}, failing to capture the finer pose and structure of objects.
%

3D lines are suitable for anchor-based representations\cite{garnett3DLaneNetEndtoEnd3D2019a, chenPersFormer3DLane2022a, liuPETRv2UnifiedFramework}, resulting in equally spaced longitudinal lines and anchor-based predictions $\Delta_{\text{xz}}$. Anchor representations reduce the representation dimension, but also restrict the degrees of freedom, limiting the maximum model capacity, which has led some works \cite{chenPersFormer3DLane2022a,huangAnchor3DLaneLearningRegress2023a} focus on more sophisticated anchor representations. 
Recent DETR-based methods\cite{baiCurveFormer3DLane2023a, luoLATR3DLane2023a, liuPETRv2UnifiedFramework} use anchor coordinates as queries, exploring the full potential of anchors. 
Unfortunately, since it's not trivial and stable to directly predict parameter-driven curves\cite{fengRethinkingEfficientLane2023, hanDecouplingCurveModeling2023}, mainstream methods stick to predicting anchor-based offsets $\Delta_\text{xz}$, 
and nearly no one ever worked directly on the raw lane format.

\subsubsection{Intersection of Task-specific Representations.}
We believe that it's already been architecturally ready to unify the two (and possibly more) perception tasks, with the population and success of DETR practices\cite{wangAnchorDETRQuery2022, liBEVFormerLearningBird2022, liuPETRv2UnifiedFramework, huangFULLERUnifiedMultimodality2023}. However, they still rely on multi-head design due to the discrepancy in task representation, which consumes more computational resources. Despite the significant differences in the dynamics and scales of the described targets, these task-specific representations only encapsulate descriptions of shapes, structures, and directions (object heading and lane ordinality). It is plausible to represent these task targets universally, taking into account perception hierarchies.

\subsection{The Task-Agnostic Representation through RepVF}
\label{sec:repvf-definition}

To facilitate efficient design in multitasking frameworks, we introduce \textit{RepVF}, a vector fields representation inspired by 2D point set object representation \cite{maRPTLearningPoint2020a, maRPTLearningPoint2020a}. Distinct from conventional isotropic points, \textit{RepVF} assigns a vector to each location making itself anisotropic. 
We define \textit{RepVF} as a collection of vector fields, each field denoted by $R_i$ as a perception target, whose semantic vector assignment mapping $\mathcal{F}_i$ is defined on $D=\bigcup_{i=1}^{n_s} S_i\subseteq \mathbb{R}^3$:
\begin{equation}
    \{R_i\},\quad R_i=(S_i,\mathcal F_i(S_i)) \subseteq \mathbb{R}^{3+d}, \quad \mathcal{F}: D \to \mathbb{R}^{d}
\end{equation}
where $d$ is the dimension of the target vector domain. Therefore, for any point $x$ within $D$, $\mathcal{F}$ assigns it a vector $\mathcal{F}(x) \in \mathbb{R}^{d}$ that encapsulates dimensional and positional attributes of the point. 
This definition allows us to cover the whole space: 
The sampling domain $D$ is partitioned into $n_s$ disjoint subsets $S_i$, with each subset containing $n$ representative sampling points: $S_i = \{(x_k,y_k,z_k)\}_{k=1}^{n}$. Accordingly, the collective vector field $\mathcal{F}$ sprawls over $D=\bigcup_{i=1}^{n_s} S_i$, articulated into $n_s$ sub-vector fields $\mathcal{F}_i$ catering to specific tasks.

Intuitively, $S_i$ directly outlines entity-level geometric structures, and $\mathcal{F}_i(S_i)$ delineates contextual characteristics (\ie $\theta$ for bounding box $B$, or adjacent point directions for lane $L$). Considering the directionality, we use \textbf{unit-length} vectors parallel to the x-y plane to capture angular behavior, resulting in behavior akin to scalars and $d=1$. To represent proper $\mathcal F_i$ for 3d lanes, directions of the penultimate points are employed for terminal lane line points.

\subsubsection{Converting RepVF into Task-Specific Representation.}
\label{sec:rep_conversion}
To leverage existing task labels for training and evaluation, similar to object detection representations based on point sets\cite{maRPTLearningPoint2020a, yangRepPointsPointSet2019}, we convert predictions \textit{RepVF} $\hat{R}_i$, category $\hat{\mathcal C}_i$ to task-specific representations. We define \textbf{geometric interpretation} processes as differentiable, parameter-free transformation functions:
\begin{equation}
    \mathcal{T}^{(\cdot)}: \hat{R}_i \rightarrow \hat{P}_i, \quad\text{\ s.t.\ } \hat{P}_i\in\{\hat{B}, \hat{L}\}_i 
\end{equation}
Specifically, we have $\mathcal{T}^{l}:\hat{R}_i\rightarrow \hat{L}_i$ for 3d lanes and $\mathcal{T}^{b}: \hat{R}_i\rightarrow \hat{B}_i$ for 3d objects. For \textbf{semantic parts}, to align $\hat{\mathcal C}_i$ with task-specific classes, we split it into $n_{(\cdot)}$ bins and take max score within each bin, where $n_{(\cdot)}$ is the count of task categories. $\hat{R}_i$ and $\hat{R}_j$ remain formally equivalent until contextualized into $\hat{P}_i$ and $\hat{P}_j$, thus maintaining task agnosticism. 

\subsubsection{Conversion Function Choice.} In our implementation, $\mathcal{T}^{l}$ sorts the unordered points of $\hat{S}_i$ by the road direction(\textit{+X} in Waymo\cite{sunScalabilityPerceptionAutonomous}). Unlike 2D boxes, 3D boxes are not strictly aligned with to \textit{XYZ} axes due to headings. Therefore,  $\mathcal{T}^{b}$ consists \textit{three parts} that respectively compute heading, box center and dimensions: 
\begin{equation}
    \mathcal{T}^b_h : \hat{\mathcal{F}}_i \rightarrow \hat{\theta_i}; \quad \mathcal{T}_{xyz}^b: \hat{S}_i \rightarrow (\hat x_i, \hat y_i, \hat z_i); \quad \mathcal{T}^b_{lwh}: \hat S_i \rightarrow (\hat l_i, \hat w_i, \hat h_i)
\end{equation}
Here, $\mathcal T^b_h$ and $\mathcal T^b_{xyz}$ are simple arithmetic averages. With $\hat \theta_i$ given by $\mathcal T^b_h$, by projecting $\hat S_i$ to the main orientation axis $\hat \theta_i$, the predictions are aligned with the predicted directions. We then proceed with two candidates of $\mathcal T^b_{lwh}$ similar to 2D practices~\cite{yangRepPointsPointSet2019}: \emph{(1)} $\mathcal{T}^b_{lwh} = \mathcal{T}_1$ , the \textbf{min-max function}, performing min/max operations projection dimensions for dimensions;\emph{(2)} $\mathcal{T}^b_{lwh} = \mathcal{T}_2$ employs a \textbf{momentum-based function} taking the variance of predicted points as the boundary box dimensions. In our practice, $\mathcal T_2$ works slightly better.

\subsubsection{Learning RepVF.}

RepVF is supervised by loss functions universally across tasks without special losses. Here we use $\text{L}^1$ regression loss on $S_i$ and $\mathcal F_i$ , and focal loss\cite{linFocalLossDense2017} on $\mathcal C_i$. Smaller structures are described by points in $S_i$, and $R_i$ itself is set-level, extending RepVF across different perception hierarchies. 
To capture this cross-level property, we adopt DETR-based structure\cite{wangAnchorDETRQuery2022, wangDETR3D3DObject2022, liuPETRPositionEmbedding2022a} and aligned query design to use raw representations motivated by \cite{liuDABDETRDYNAMICANCHOR2022a, baiCurveFormer3DLane2023a}.
With multiple decoder layers\cite{zhuDeformableDETRDeformable2021d, wangDETR3D3DObject2022}, the learning process of RepVF can be characterized as \textit{iterative} update $\Omega_l$ on queries $Q_{l}$ and inferring $\hat{R}_l, \hat{\mathcal C}_l$ from them with linear layer $\Phi$ and decoder layer $l$:
\begin{equation}
    Q_{l} = \Omega_l(Q_{l-1}),\quad \hat{R}_{l}, \hat{\mathcal C}_l=\Phi(Q_{l})
\end{equation}
We will explain our network in detail in the next section.

%% file: sec/4_RFTR.tex
\begin{figure}[tb]
    \centering
    \includegraphics[width=\textwidth]{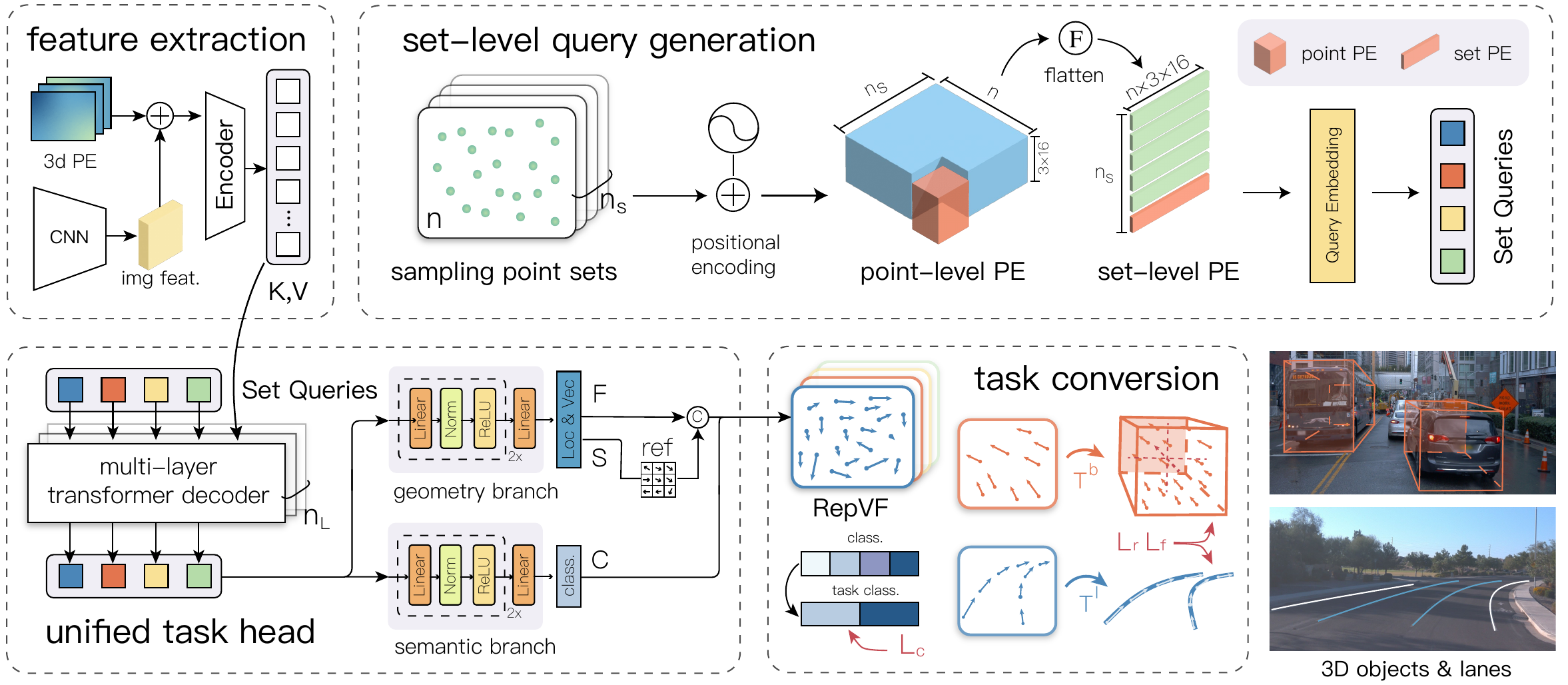}
    \caption{
    Overview of our RFTR (Representative Vector Fields Transformer) built upon \textit{RepVF}. Multi-view image features with 3D position embeds are extracted by image backbone and fed into the decoder. We generate set-level query embeds from 3D space sampling sets, each representing a perception target. One single unified task head is then used to predict unified representative vector fields. Finally, predictions are transformed in a differentiable manner into task specific representations to utilize existing labels for supervision.
    }
    \label{fig:framework}
\end{figure}

\section{RFTR: The Single-head Multi-tasker}

In this section, we introduce the single-head multi-task framework RFTR (Representative Vector Fields Transformer), which uses \textit{RepVF} representation to unify different tasks and exploit intrinsic connections among them. We also observed one interesting behavior of RFTR in the balance of multi-task gradients.
RFTR is built upon DETR networks\cite{carionEndtoEndObjectDetection2020a, wangDETR3D3DObject2022, liuPETRPositionEmbedding2022a} and is end-to-end on all tasks.

\subsection{Network Structure}
\Cref{fig:framework} shows our RFTR architecture built upon \textit{RepVF}. It inputs images $I$ from $N$ cameras with known camera parameters, and outputs unified representative vector fields $\mathcal F$ for various perception tasks in the 3D scene. RFTR contains four main components that extract features from images, generate set-level queries for hierarchical representation structure, solve unified representative vector fields and multiple tasks with one single head, and convert results to task representations in a differentiable way to utilize existing task labels.

\subsubsection{Feature Extraction.} Following common practices in utilizing 2D backbones to extract features\cite{wangDETR3D3DObject2022, liuPETRPositionEmbedding2022a, liBEVFormerLearningBird2022}, we adopt ResNet\cite{heDeepResidualLearning2016} (ResNet-50) and FPN\cite{linFeaturePyramidNetworks2017} to produce 2D features $F^{2d}$. They're encoded with 3D position embedding in PETR\cite{liuPETRPositionEmbedding2022a} into 3D position aware features $F^{3d}$.

\subsubsection{Generating Set-level Perception Queries.}

As \textit{RepVF} is sampled from $n_s$ point sets $S_i$ with $n$ points each and is two-level, our aim is to generate queries that are consistent with this hierarchical point-set structure.
Hence provide the decoder with sufficient spatial information, improving feature interactions.
Unlike former approaches~\cite{wangDETR3D3DObject2022, luoLATR3DLane2023a, liuPETRPositionEmbedding2022a, liuPETRv2UnifiedFramework} that encode one single 3D location, each query in RFTR encodes multiple locations simultaneously. 

We generate set-level queries $Q_0\in\mathbb{R}^{n_s\times d_q}$ from initial point sets corresponding to $S_i$, where $d_q$ is the dimension of each query. The generation process $\mathcal G$ consists of encoding initialized positions from the point level, and embedding them at the set level: 
\begin{equation}
    \mathcal G = \text{embed}(\text{Flatten}(\text{PE}))
\end{equation}
where PE is the positional encoding\cite{vaswaniAttentionAllYou2017a}. Point-level PE is flattened into set-level and embedded by a small MLP with two layers. 
To avoid confusing the model with expansive locations falling on multiple targets in the early stages of training, we use $n$ identical 3D locations sampled with 0-1 uniform distribution for all $n_s$ initial point sets.

To our best knowledge, the closest generation approach to ours is \cite{baiCurveFormer3DLane2023a}, which generates query from an ordered sequence of anchor points motivated by \cite{liuDABDETRDYNAMICANCHOR2022a}. However, ours is from an unordered sampling point set and is not task-specific. Moreover, we use a tiny 16-dimensional PE for each coordination, divert from former methods \cite{wangDETR3D3DObject2022, liuPETRPositionEmbedding2022a} with a much larger 256-dimension. In our multi-task setting, queries are equivalently initialized, thus task-agnostic. We divide the first 80\% as 3D objects and the last 20\% as 3D lanes.

\subsubsection{Single-head Multi-tasking.}
Similar to all previous approaches\cite{carionEndtoEndObjectDetection2020a, wangDETR3D3DObject2022, liuPETRPositionEmbedding2022a}, queries are learnable and updated in decoder layers \textit{iteratively}. The unified head $\Phi$ of RFTR is composed by a multi-layer transformer decoder\cite{vaswaniAttentionAllYou2017a}, and two branches $\Phi_g$ and $\Phi_s$ dedicated for geometric regression and semantic prediction respectively to predict \textit{RepVF}. We use the same $n_L=6$ decoder layers with previous methods\cite{carionEndtoEndObjectDetection2020a, wangDETR3D3DObject2022, liuPETRPositionEmbedding2022a} to exploit intrinsic tasks connections implicitly. The interaction process within these layers can be formally expressed as:
\begin{equation}
    Q_l=\Omega_l(F^{3d},Q_{l-1}),\quad l=1,\dots,N_L
\end{equation}
With unified head $\Phi=\{\Phi_g, \Phi_s\}$,  predictions $\hat{R}_{li}$ and $\hat{\mathcal C}_{li}$ is decoded from $i$-th query at $l$-th layer:
\begin{equation}
    \hat{R}_{li} = \Phi_g(Q_{li}), \quad \hat{\mathcal C}_{li} = \Phi_s(Q_{li})
\end{equation}
For $\Phi_g$, additional offsets are added to produce $S_{li}$. Unified geometric predictions $\hat{R}_l=(\hat{S}_l, \hat{F}_l) \subset \mathbb R^{n_s\times(3+d_p)}$ are converted to task-specific representations:
\begin{equation}
    \hat{L}_{li} = \mathcal{T}^l(\hat{R}_{li}), \quad \hat{B}_{li} = (\mathcal T^b_{xyz}, \mathcal T^b_{\theta}, \mathcal T^b_{lwh})(\hat{R}_{li})
\end{equation}
Here,  $\mathcal T^{(\cdot)}$ are differentiable functions predefined in section \ref{sec:rep_conversion}. Each head branch $\Phi_{(\cdot)}$is a simple small MLP network with three fully connected layers. Following DETR3D\cite{wangDETR3D3DObject2022}, we weight the loss for the predictions from each decoder layer $\Omega_l$ during training, and only outputs from the last layer is used during inference. 

\subsubsection{Loss Function.}
RFTR uses existing task labels for supervision, and its loss is calculated between converted predictions and task ground truths. It consists of the following components for coordinate regression, vector fields regression and classification respectively:
\begin{equation}
\mathcal{L}^{(\cdot)}=
w_r^{(\cdot)}\mathcal{L}_{r}^{(\cdot)}+
w_f^{(\cdot)}\mathcal{L}_{f}^{(\cdot)}+
w_c^{(\cdot)}\mathcal{L}_{c}^{(\cdot)},\quad 
\mathcal{L}=\sum_{\text{tasks}}{\mathcal{L}^{(\cdot)}}
\end{equation}
where $w^{(\cdot)}$ represent loss component weights for task ${(\cdot)}$, \ie 3D object detection and 3D lane detection. We use the Hungarian algorithm\cite{kuhnHungarianMethodAssignment1955} for label assignment after cost calculation between ground truth and converted predictions $\hat{B}_{li}, \hat{P}_{li}$. 

In practice, we choose $w^{b}_c=w^{l}_c=2.0$, $w^{l}_r=0.003$, $w^{b}_r=0.1$, $w^{b}_f=0.2$, and $w^l_f=0.032$. 
Coordinate regression and classification weights are selected empirically to equalize the scales and contributions of each loss term \cite{zhangFairMOTFairnessDetection2021}, fields weights are empirically selected. 
We use $\text{L}^1$ loss for $R_i$ regression, and focal loss \cite{linFocalLossDense2017} with $\gamma=2.0$ and $\alpha=0.25$ for classification. Since we predict a degenerated scalar field, we supervise the angular prediction by its sine and cosine values.
All tasks share the same losses and similar weights, no additional, task-specific loss measures are adopted.

\begin{figure}[tb]
  \centering
  \resizebox{0.9\textwidth}{!}{
  \begin{subfigure}{0.60\linewidth}
    \includegraphics[width=\textwidth]{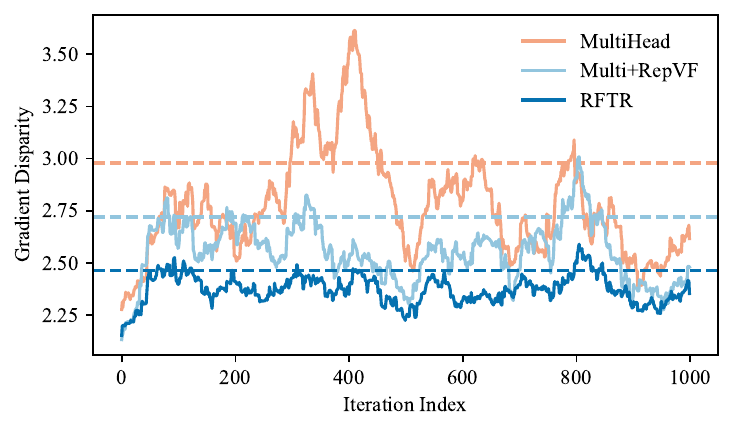}
    \caption{Gradient disparities by iteration index.}
    \label{fig:gradient-compare-a}
  \end{subfigure}
  \hfill
  \begin{subfigure}{0.36\linewidth}
    \includegraphics[width=\textwidth]{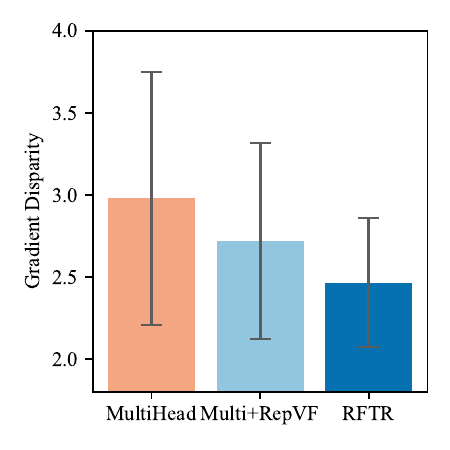}
    \caption{Average gradient disparities.}
    \label{fig:gradient-compare-b}
  \end{subfigure}
  }
  \caption{Our single-head RFTR model shows a reduced gradient imbalance and better disparity stability (mean: 2.47, variance: 0.90) compared to the multi-head baseline (mean: 2.98 variance: 1.86) or the multi-head baseline with RepVF (mean: 2.72, variance: 1.57).
Ideally balanced gradient disparity through training iterations should approach 2. 
  For ease of trend observation, we have clipped the top 2\% values to the mean, with curves smoothed through a window size of 50 1-d convolution.}
  \label{fig:gradient-compare}
\end{figure}

\subsection{Gradient Balance in Single-head Multi-tasking}

The multi-head multi-task approach often leads to task conflict, attributed in prior studies\cite{navonMultiTaskLearningBargaining2022, liuImpartialMultitaskLearning2020, 
kendallMultiTaskLearningUsing2018, senerMultiTaskLearningMultiObjective2018} to gradient imbalance. 
These studies\cite{liuImpartialMultitaskLearning2020, navonMultiTaskLearningBargaining2022} and more recent work\cite{huangFULLERUnifiedMultimodality2023} have explored appropriate gradient calibration strategies to address this issue.
As we will show, our single-head RFTR naturally achieves better balanced multi-task gradients thus calibration may no longer be required.

To elucidate the advantage of our single-head multi-task paradigm over multi-head multi-task settings, we use an improved measure of \textbf{gradient disparity} adapted from FULLER\cite{huangFULLERUnifiedMultimodality2023}, by taking into account the symmetry disparity:
\begin{equation}
    \text{diff}(\nabla \mathcal{L}^{l}, \nabla \mathcal{L}^b)=
    \frac{\Vert\nabla \mathcal{L}^l\Vert}{\Vert\nabla \mathcal{L}^b\Vert}+
    \frac{\Vert\nabla \mathcal{L}^b\Vert}{\Vert\nabla \mathcal{L}^l\Vert}
\end{equation}
where $\Vert \cdot \Vert$ is the Frobenius norm and $\nabla \mathcal L^{(\cdot)}$ denotes gradients. Given that $\Vert\nabla \mathcal L^{(\cdot)}\Vert$ is always greater than 0, the value of $\textit{diff}$ function is always greater than or equal to 2. Therefore, the more stable and closer to 2 the gradient disparity is during model training, the better the multi-task stability indicated.

\Cref{fig:gradient-compare-a} illustrates the gradient disparity curves through training iterations. The multi-head multi-task method\cite{liuPETRPositionEmbedding2022a, liuPETRv2UnifiedFramework} is under the same task settings with ours.  In the sampling 1k iterations, the stability and absolute scale of the gradient disparity curve for the multi-head configuration are inferior to our single-head multi-task structure. The mean and variance of the multi-head disparities are about 108.5\% and 107.8\% higher, respectively, than our single-headed RFTR approach, as shown in \cref{fig:gradient-compare-b}. We have also observed that unified representation and single head design contributed to the improvement about equally.
More importantly, this means that our single-head RFTR \textit{mitigates}, though not eliminating, the gradient imbalance between tasks, which is fundamentally different from task gradient balancing strategies that do not change the gradient itself.

%% file: sec/5_experiments.tex
\section{Experiments}

We evaluate our method on the Waymo Open Dataset \cite{sunScalabilityPerceptionAutonomous} extended by 3D lanes annotated in OpenLane \cite{chenPersFormer3DLane2022a}, so that we train and evaluate tasks simultaneously. 

\subsection{Datasets}
\subsubsection{Waymo Open Dataset and OpenLane.} Waymo Open Dataset (Perception) \cite{sunScalabilityPerceptionAutonomous} is a large-scale multimodal dataset composed of data from 5 cameras(front and sides), 1 mid-range lidar, and 4 short-range lidars. It contains 1000 sequences in total(about 198k samples), a training set of 798 sequences and a validation set of 202 sequences. 
OpenLane \cite{chenPersFormer3DLane2022a} is a comprehensive real-world 3D lane detection dataset built upon the Waymo Open dataset \cite{sunScalabilityPerceptionAutonomous}. It includes a vast array of 200K frames and over 880K carefully annotated lanes, making it one of the largest of its kind to date. The dataset encompasses complex lane structures and features a variety of scene tags, such as weather and locations. Designed to emulate real-world scenarios, OpenLane provides a challenging benchmark for advanced lane detection algorithms. We integrate its 3D lane labels with the original Waymo Open sensor data for multi-task annotation.

\subsubsection{Alignment of Datasets.}  To achieve both tasks simultaneously, we align both datasets in terms of coordinate systems and data splits. 
\textit{(1)} Coordinate systems. For the vehicle frame, OpenLane \cite{chenPersFormer3DLane2022a} uses a vehicle coordinate system that corresponds to the "y-front, x-right, z-up" positioning of 3D-LaneNet \cite{garnett3DLaneNetEndtoEnd3D2019a}. We then align this system to coincide with the commonly utilized "x-front, y-right, z-up" axis for lidar-based 3D object detection. With respect to the sensor frame, we apply transformations to the camera intrinsics so that the sensor frames for all cameras conform to a uniform "z-depth, x-right, y-down" frame. The predictions are then transformed back into their respective task-based coordinate systems for evaluation.  
\textit{(2)} Data splitting. The OpenLane \cite{chenPersFormer3DLane2022a} dataset provides a 1000 segment full version and a 300 segment smaller subset (30\%). The partitioning of the larger version complements the original train/validation split of the Waymo Open Dataset. The subset version partitioning is aligned with OpenLane. We perform ablation studies on the smaller subset.

\input{sec/tables/openlane}

\input{sec/tables/waymo}

\begin{figure*}[tbp]
    \centering
     \includegraphics[width=\textwidth]{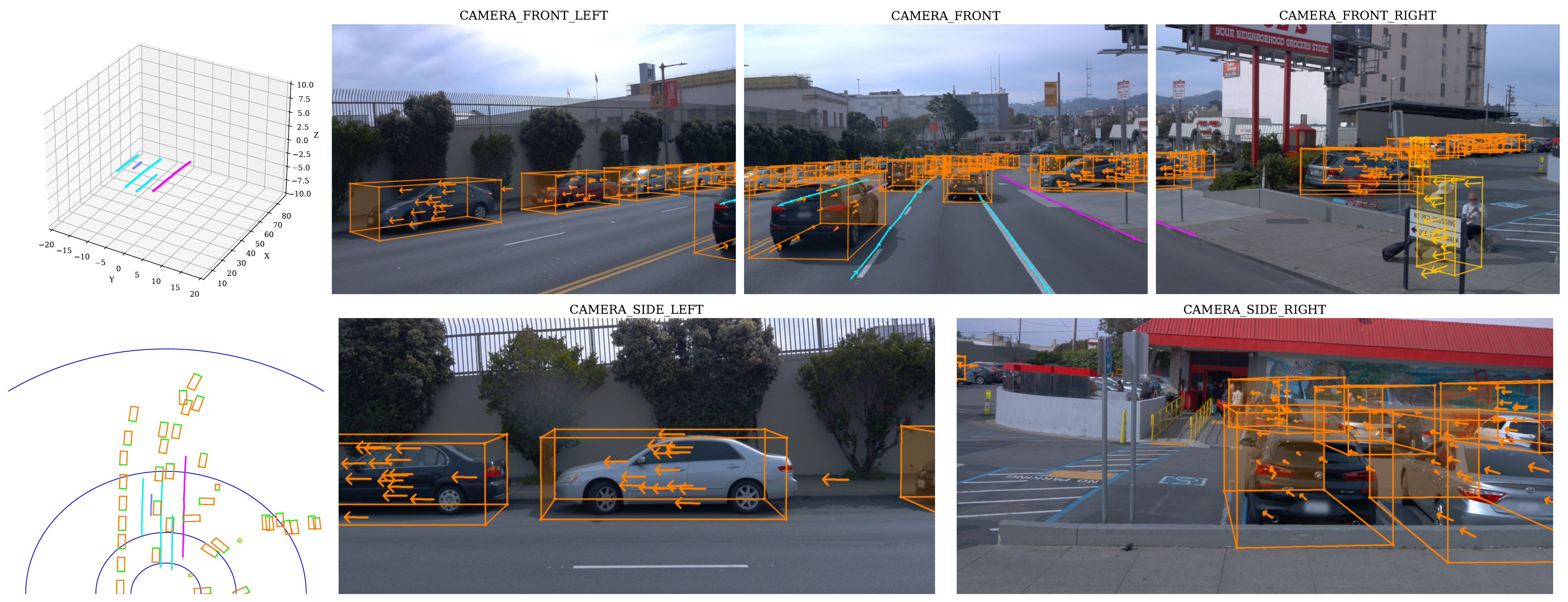}

    \caption{Visualization of the RFTR model's performance and the learned \textit{RepVF}. 
    }
    \label{fig:qualitative}
\end{figure*}

\subsection{Implementation Details}
\subsubsection{Data Processing.} We use the v1.4.2 of Waymo Open Dataset \cite{sunScalabilityPerceptionAutonomous}, and v1.2 for OpenLane \cite{chenPersFormer3DLane2022a}. We set the perception range to $[-84.88m, 84.88m]$ for the X and Y axis, and $[-10m, 10m]$ for the Z axis. 3D objects that do not fall within the perception range, are invisible to the cameras, or do not have LiDAR points, are filtered. 
For 3D lanes, visibility filtering is first applied after axis transformation, then 3D lanes outside the perception range are pruned and then resampled to a fixed size. Approximately 2.07\% of the frames without annotations after filtering are excluded from use. 

\subsubsection{Model and Training.} We used ResNet 50~ \cite{heDeepResidualLearning2016}  as the 2D image backbone for fair comparisons. Following PETR~\cite{liuPETRPositionEmbedding2022a}, the same upsampled and fused P4 feature with 1/16 input resolution is used as the 2D feature. A CPFPN~\cite{linFeaturePyramidNetworks2017} is used as the image neck. All 3D coordinates are normalized to $[0,1]$. The same weights are used for both the loss computation and the bipartite matching. We adopt progressive weighting for the loss of each decoder layer to improve the convergence speed. For 3D object detection, we compute matching costs based on the center displacement and class score weightings. Dimensions are refined post-matching using the ground truth headings to increase training robustness. We use a total of $n_s=750$ queries with $n=15$ sample locations each for the two tasks, 600 for 3D objects and 150 for 3D lanes. 
We trained our model using the AdamW~ \cite{loshchilovDecoupledWeightDecay2019}  optimizer with a 0.01 weight decay. The learning rate started at $2.0\times10^{-4}$ and was decayed with a cosine annealing policy~\cite{loshchilovSGDRStochasticGradient2016}. We adopted a multi-scale training strategy, with the shorter side chosen randomly from $[640,900]$ and the longer side not exceeding 1600. We applied rotation augmentations between $[-\pi/18,\pi/18]$. We resized images of sizes $1920\times1080$ and $1920\times 1280$ from different views to $1600 \times 900$ before augmenting. The final resolution for network inputs was $1408\times512$. All experiments on the full dataset are trained for 24 epochs on 4 GeForce RTX 4090 GPUs with a batch size of 8. No test time augmentation methods are used during inference.

\subsection{Comparison with SOTA}

\subsubsection{3D Lane Detection.} As shown in \cref{tab:openlane_1000}, our RFTR model showcases significant achievements in 3D lane detection on the OpenLane validation set, demonstrating its competitive edge with a high F-score of 61.8 and category accuracy of 91.6\%. It matches or exceeds state-of-the-art models such as PETRv2-V-400\cite{liuPETRv2UnifiedFramework} (more than $26\times$ denser representation than ours) when assessed on the OpenLane \cite{chenPersFormer3DLane2022a} validation set. Notably, while our model has a slightly higher X error in the "far" class than \cite{huangAnchor3DLaneLearningRegress2023a}, it manages to lead with a higher F-score and near accurate Z errors. 
Moreover, our method yields notable performance improvements over the Persformer \cite{chenPersFormer3DLane2022a} baseline, reflecting the advantageous adaptation of our unified \textit{RepVF} representation and the RFTR architecture.

\subsubsection{3D Object Detection.}
The performance of our RFTR model in 3D object detection, as demonstrated in \cref{tab:waymo_1000}, shows comparable promising results on the Waymo Open Dataset \cite{sunScalabilityPerceptionAutonomous} validation set. 
While our approach shows a commanding lead in detecting vehicles and cyclists with a solid L1 mAP and mAPH, it shows challenges in detecting smaller entities such as pedestrians. It should be noted, however, that these preliminary results indicate the great potential of using our unified perceptual model for complex tasks, which will require further fine-tuning for object classes with less prominent features.

\subsection{Ablation Studies}
\subsubsection{Head design.} We begin by finding out how head designs and representations contribute to the performance, in \cref{tab:new-multi}. The \textit{RepVF} representation with the single-head design, \ie RFTR, performs the best in general with 37.4M parameters. Unified representation brings performance improvements in multi-task learning and reduces task competition. In contrast, task-specific multi-head design produces inferior and unstably biased performance due to task competition, even with more parameters.
\subsubsection{Representation.} Moreover, we have evaluated the effectiveness of different representations for 3D lane detection in \cref{tab:ablation_multi-tasking}, including traditional anchor representations\cite{liuPETRPositionEmbedding2022a,chenPersFormer3DLane2022a}, simple randomly sampled point set representation (PS), and our proposed representative vector fields (RepVF).
This study reveals that \textit{RepVF} significantly outperforms traditional anchor and PS methods, and incorporating multi-task learning marginally improves these metrics.
\input{sec/tables/ablations}

\subsection{Qualitative Results}

\Cref{fig:qualitative} demonstrates the capability of RFTR to simultaneously achieve 3D lane detection and 3D object detection on the extended \cite{liVisualBERTSimplePerformant2019} Waymo open dataset\cite{sunScalabilityPerceptionAutonomous}. The figure clearly illustrates that RFTR can accurately predict \textit{RepVF} and perform both tasks at once. In terms of 3D lane detection, RFTR is capable of predicting the shape of lane lines in areas obscured by vehicles, demonstrating its robustness against occlusion. Regarding 3D object detection, the unified RepVF representation accurately captures the 3D bounding boxes. This dual achievement highlights RFTR's effectiveness in handling complex driving scenarios, enhancing both navigation and safety features in autonomous driving systems.

%% file: sec/tables/openlane.tex
\newcommand{\red}[1]{\textcolor{red}{#1}}
\newcommand{\blu}[1]{\textcolor{blue}{#1}}
\begin{table*}[tb]
    \caption{Comparison of recent models on 3D lane detection using the OpenLane \cite{chenPersFormer3DLane2022a} validation set. 
    The down arrow
    indicates that lower metric values correspond to better performance, and vice versa.
    Red indicates the best result, and blue the second-best.
    }
    \label{tab:openlane_1000}
    \centering
    \resizebox{0.9\textwidth}{!}{
    \setlength{\tabcolsep}{4mm}{
    \begin{tabular}{l|clcccc}
    \toprule
        \multicolumn{1}{l}{} &  & Category  & \multicolumn{2}{c}{X error (m)} $\downarrow$ &  \multicolumn{2}{c}{Z error (m) $\downarrow$ }\\
        \cmidrule{4-5} \cmidrule{6-7}
        \multicolumn{1}{c}{\multirow{-2}{*}{Methods}}& \multirow{-2}{*}{F-Score $\uparrow$} & Accuracy \multirow{-2}{*}{$\uparrow$} & \textit{near} & \textit{far} & \textit{near} & \textit{far} \\
    \midrule
    \hline
 3D-LaneNet\cite{garnett3DLaneNetEndtoEnd3D2019a}~\tiny{[ICCV19]}&  44.1& -& 0.479& 0.572& 0.367& 0.443\\
 Gen-LaneNet\cite{guoGenLaneNetGeneralizedScalable2020}~\tiny{[ECCV20]}& 32.3& -& 0.591& 0.684& 0.411&0.521\\
  PersFormer\cite{chenPersFormer3DLane2022a}~\tiny{[ECCV22]}& 50.5& \red{92.3}& 0.485& 0.553& 0.364&0.431\\
   Cond-IPM\cite{chenPersFormer3DLane2022a}& 36.6& -& 0.563& 1.080& 0.421&0.892\\
 CurveFormer\cite{baiCurveFormer3DLane2023a}~\tiny{[ICRA23]}& 50.5& -& 0.340& 0.772& 0.207&0.651\\
 PETRv2-E\cite{liuPETRv2UnifiedFramework}~\tiny{[ICCV23]}& 51.9& -& 0.493& 0.643& 0.322&0.463\\
PETRv2-V-10\cite{liuPETRv2UnifiedFramework}& 57.8 & -& 0.427& 0.582& 0.293&0.421\\
 PETRv2-V-400\cite{liuPETRv2UnifiedFramework}& \blu{61.2}& -& 0.400& 0.573& 0.265&0.413\\
 BEV-LaneDet\cite{wangBEVLaneDetEfficient3D2023}~\tiny{[CVPR23]}& 58.4& -& 0.309& 0.659& 0.244&0.631\\
 Anchor3DLane\cite{huangAnchor3DLaneLearningRegress2023a}~\tiny{[CVPR23]}& 53.1& 90.0& \red{0.300}& \red{0.311}& \blu{0.103} & \blu{0.139}\\
SPG\cite{yaoSparsePointGuided}~\tiny{[ICCV23]} & 52.3 & - & 0.468 & 0.514 & 0.371 & 0.418 \\

    \midrule
RFTR (Ours) & \red{61.8} & \blu{91.6} & \blu{0.341} & \blu{0.450} & \red{0.073} & \red{0.107} \\
    \bottomrule
    \end{tabular}}}

\end{table*}

%% file: sec/tables/waymo.tex
\begin{table*}[tb]
    \caption{Comparison of recent works of 3D object detection on the Waymo Open Dataset \cite{sunScalabilityPerceptionAutonomous} val set. 
    mAPs are LEVEL\_1 in Waymo metrics.}
    \label{tab:waymo_1000}
    \centering
    \resizebox{0.9\textwidth}{!}{
    \setlength{\tabcolsep}{3mm}{
    \begin{tabular}{l|cccccccc}
    \toprule
 & \multicolumn{2}{c}{Overall}& \multicolumn{2}{c}{Vehicle}& \multicolumn{2}{c}{Cyclist}& \multicolumn{2}{c}{Pedestrian}\\
         \multirow{-2}{*}{Methods}&  mAP& mAPH & mAP& mAPH  & mAP& mAPH  &mAP&mAPH  \\
         \midrule
         \hline
         DETR3D\cite{wangDETR3D3DObject2022} & 10.1 & 9.6 & 9.2 & 9.1 & 15.8 & 15.5 & 5.3 & 4.1 \\ 
         PETR\cite{liuPETRPositionEmbedding2022a}& 20.9 & 19.7 & 31.1 & 30.8 & 19.5 & 17.6 & 12.1& 10.6\\
         \midrule
RFTR (Ours) & 19.5 & 18.7 & 22.6 & 22.4 & 27.8 & 26.3 & 8.0 & 7.2  \\
  \bottomrule
    \end{tabular}}}
\end{table*}

%% file: sec/tables/ablations.tex
\begin{table}[t]
\begin{minipage}{0.54\textwidth}
        
        \centering
        \caption{Multi-task head design and representation ablation. Tasks are trained jointly on 30\% data of Waymo Open Dataset (L1 mAP) and OpenLane (F-Score). TS denotes task-specific representation. $^\dagger$: lightweight models. }
        \label{tab:new-multi}
        {\small
    \resizebox{0.9\linewidth}{!}{
            \begin{tabular}{cc|c|c|c|c}
            \toprule
                RepVF&TS&Head&Params$\downarrow$&  F-Score$\uparrow$&Vehicle$\uparrow$\\
             \midrule
     & \checkmark& Multi $^\dagger$& 37.1M&  49.6& 20.1\\
     &\checkmark& Multi&46.7M&  58.7&17.8\\
     \checkmark& & Multi$^\dagger$&39.7M&  63.4& 18.2\\
     \checkmark&&Multi&49.2M&  66.4&22.1\\
             \midrule
     \checkmark&&Single&37.4M&  66.5&25.3\\
              \bottomrule
        \end{tabular}}
    }

\end{minipage}%
\hfill
\begin{minipage}{0.44\textwidth}
        \caption{Results with different representations on 3D lane. 'Acc' for category accuracy and 'Error' for X error near. MT means simultaneous multi-task 3D lane and 3D object detection. }
        \label{tab:ablation_multi-tasking}
        \centering
          \resizebox{0.95\textwidth}{!}{
          \begin{tabular}{c|ccc}
    \toprule
       Representation& F-Score$\uparrow$ & Acc$\uparrow$ & Error$\downarrow$\\
    \midrule
       Anchor & 53.4 & 86.0 & 0.476 \\
        PS & 58.1 & 88.0 & 0.545 \\
         RepVF & 66.0 & 91.4 & 0.439 \\
         \midrule
         RepVF + MT& 66.5 & 91.7 & 0.420 \\
    \bottomrule
    \end{tabular}
        }

\end{minipage}
\end{table}